\documentclass[conference]{IEEEtran}
\IEEEoverridecommandlockouts
\usepackage{cite}
\usepackage{amsmath,amssymb,amsfonts}
\usepackage{graphicx}
\usepackage{textcomp}
\usepackage{xcolor}
\def\BibTeX{{\rm B\kern-.05em{\sc i\kern-.025em b}\kern-.08em
    T\kern-.1667em\lower.7ex\hbox{E}\kern-.125emX}}
    
\usepackage{graphicx}
\usepackage{amsfonts}
\usepackage{amsmath}
\usepackage{boldline}
\usepackage{amsthm}
\usepackage{epstopdf}
\usepackage{amssymb}
\usepackage{tabu}
\usepackage{multirow}
\usepackage{multicol}
\usepackage{booktabs}
\usepackage{setspace}
\usepackage{algpseudocode}
\usepackage{algcompatible}
\usepackage{mathrsfs,dsfont}
\usepackage{color}
\usepackage{float}
\usepackage{enumitem}
\usepackage{epsfig,cite}
\usepackage{url}
\usepackage{cases}
\usepackage{mathtools}
\usepackage{amsfonts}
\usepackage{nccmath}
\usepackage{nomencl}
\usepackage{etoolbox}
\makenomenclature
\renewcommand\nomgroup[1]{%
  \item[\bfseries
  \ifstrequal{#1}{I}{Sets and Indices}{%
  \ifstrequal{#1}{P}{Parameters}{%
  \ifstrequal{#1}{V}{Variables}{}}}%
]}

\usepackage{stackengine}
\usepackage[caption = false]{subfig}
\usepackage[linesnumbered,lined,boxed,commentsnumbered,ruled,longend]{algorithm2e}
\newcommand{\tabincell}[2]{\begin{tabular}{@{}#1@{}}#2\end{tabular}}
\usepackage{varwidth}
\usepackage{tikz}
\usetikzlibrary{arrows}
\usepackage{fancyhdr}

\usetikzlibrary{shapes, arrows,positioning}
\usepackage{tikz}
\usetikzlibrary{shapes,arrows.meta,decorations}
\tikzset{block/.style={draw, fill=blue!15, rectangle, minimum height=1.7em, minimum width=2em},
sum/.style={draw, fill=blue!15, circle, node distance=1cm},
pmu/.style={draw, fill=blue!15, circle, node distance=1cm},
blkk/.style={draw, fill=blue!15, rectangle, minimum height=0.8em, minimum width=1em},
blck/.style={draw, fill=blue!15, rectangle, minimum height=0.8em, minimum width=1em},
blocklarge/.style={draw, fill=blue!15, rectangle, minimum height=2cm, minimum width=2cm},
blocklarger/.style={draw, fill=blue!15, rectangle, minimum height=2.5cm, minimum width=2.5cm},
blocksina/.style={draw, fill=blue!15, rectangle, minimum height=0.8cm, minimum width=0.8cm},
blocksinaa/.style={draw, fill=blue!15, rectangle, minimum height=0.8cm, minimum width=1.1cm},
input/.style={coordinate},
output/.style={coordinate},
dot/.style = {circle,fill, inner sep=0.01pt, node contents={}},
alr/.style = {Stealth-Stealth},
arr/.style = {-Stealth},
HVgate/.style = {signal, draw, fill=blue!15, signal to=east,
font=\small, align=left},
custom pin/.style={pin edge={solid,thick,black,decorate,font = \small, decoration={strange pin}}}}

\newcommand*{\itemizeTabular}[2][l]{%
  \begin{tabular}{!{\kern\tabcolsep\usebeamertemplate{itemize item}}#1@{}}#2\end{tabular}}
\usetikzlibrary{positioning,arrows.meta}

\begin{document}

\title{Analyzing the Travel and Charging Behavior  of Electric Vehicles  -- A Data-driven Approach
}
\author{\IEEEauthorblockN{Sina Baghali}
\IEEEauthorblockA{Civil, Environmental, and Construction\\
Engineering\\
University of Central Florida\\
Orlando, Florida, USA\\
baghalisina@knights.ucf.edu}
\and
\IEEEauthorblockN{Samiul Hasan}
\IEEEauthorblockA{Civil, Environmental, and Construction\\ Engineering\\
University of Central Florida\\
Orlando, Florida, USA\\
samiul.hasan@ucf.edu}
\and
\IEEEauthorblockN{Zhaomiao Guo}
\IEEEauthorblockA{Civil, Environmental, and Construction\\ Engineering\\
University of Central Florida\\
Orlando, Florida, USA\\
guo@ucf.edu}
}

\maketitle

\begin{abstract}
The increasing market penetration of electric vehicles (EVs) may pose significant electricity demand on power systems. This electricity demand is affected by the inherent uncertainties of EVs' travel behavior that makes forecasting the daily charging demand (CD) very challenging. In this project, we use the National House Hold Survey (NHTS) data to form sequences of trips, and develop machine learning models to predict the parameters of the next trip of the drivers, including trip start time, end time, and distance. These parameters are later used to model the temporal charging behavior of EVs. The simulation results show that the proposed modeling can effectively estimate the daily CD pattern based on travel behavior of EVs, and simple machine learning techniques can forecast the travel parameters with acceptable accuracy.
\end{abstract}

\begin{IEEEkeywords}
Electrical vehicles, travel behavior, charging demand, machine learning. 
\end{IEEEkeywords}

\section{Introduction}

\subsection{Motivation}

Growing needs for controlling the greenhouse gases emissions and fossil fuel consumption promote the development of more efficient electric Vehicles (EVs) to compete with the conventional gasoline vehicles. In addition, different countries have developed incentive programs to accelerate EV adoption, and recently, EVs are considered to improve system resilience \cite{haggi2019review}. According to the International Energy Agency (IEA), the number of EVs have passed 7.2 million in 2019 worldwide, and estimated that the EV fleet will impose  1,000 TWh of electricity demand by 2030 \cite{outlook2020electric}. The charging demand (CD) of EVs is closely related to their travel behavior and will add a tremendous load on the electricity grid. Therefore, the directed progress of the EV fleet will have significant impacts on  both  transportation  and  power  systems \cite{TTE2021}. In this paper, we present a data-driven approach to model the travel behavior of EVs by predicting the travel parameters i.e., start and end time and the travel distance of their trips. Then, these predictions are used to estimate the CD patterns of EVs.

\subsection{Literature Review}
Forecasting the CD of EVs has been a topic of interest among researchers, resulting in an extensive literature. Stochastic modeling is one of the most popular methods applied in many studies for investigating the stochastic behavior of CD \cite{ashtari2011pev, li2018gis, qian2010modeling, arias2016electric, leou2013stochastic, tang2015probabilistic, sun2017novel}. Monte-Carlo Simulation (MCS) is the basis of the stochastic modeling, where a significant number of scenarios are generated for the travel parameters and  develop trip chains to calculate the charging behavior of EVs \cite{qian2010modeling, leou2013stochastic,wang2014analysis,arias2016electric,shun2016charging}. One of the drawbacks of MCS is that MCS needs to generate a large number of scenarios to be accurate, which will make the problem computationally expensive. Moreover, the correlation of travel parameters is neglected in this method since scenarios are generated independently for each parameter. Ashtari et.al \cite{ashtari2011pev} have proposed a stochastic method to incorporate the correlation of the parameters in their distributions. However, this method only considers the correlation of two parameters. Markov-chain model is another stochastic method implemented for modeling the charging behavior of EVs \cite{sun2017novel}. This method requires less number of scenarios generated compared to the MCS. However, it needs to group the status of EVs’ traveling into different steps to form an accurate enough Markov transition matrix which still has computational challenges. In addition, aggregating EVs' traveling status in discrete steps adds inaccuracy to the model.

Machine learning algorithms, such as k-nearest neighbours (KNN) \cite{li2018gis} or artificial neural networks \cite{panahi2015forecasting,jahangir2019charging,mansour2020machine}, are recently adopted in CD estimation.  However, studies in \cite{panahi2015forecasting,mansour2020machine} incorporate probabilistic models to generate synthetic trips to overcome the small scale of their input data. In \cite{jahangir2019charging}, a large survey data is used for estimating the travel parameters and CD. However, the authors assume smart charging of EV users where a centralized entity determines the optimal time for charging and shifts the charging time to the hours with low charging price, which does not represents the realistic daily CD. Additionally, they have assumed that vehicle will charge only after their last trip of the day, ruling out the possibility of charging during work hours and other public charging options. 

In summary, on one hand, the low estimation accuracy and computation challenges are the main limitations of the probabilistic models; on the other hand, generating synthetic data and making oversimplified assumptions are the main drawbacks of current literature on CD estimation using machine learning models.

\subsection{Contribution}
In this study, we seek to resolve the mentioned drawbacks by developing machine learning models that estimate the realistic daily CD profile based on the predicted travel behavior of EVs, relaxing some of the unrealistic assumptions, e.g., EV drivers make decentralized charging decision and can charge at locations other than home. The modeling and estimations are based on the National Household Travel Survey (NHTS) \cite{NHTS2017}, and no probability distribution is used in forecasting in contrast to the mentioned studies. In addition, we investigate a variety of machine learning approaches to determine the best forecasting model. We find that random forest (RF) regressor provides the best accuracy with the least error value, and artificial neural network (ANN) based methods are able to adapt to the trend of CD during the day.

The remainder of the paper is organized as follows. Section \ref{sec:data_exploration} delineate the survey data and pre-processing steps. CD modeling is explained in Section \ref{sec:demand_model}. Forecasting models and their characteristics are included in Section \ref{sec:Models}. Section \ref{sec:results} presents the comparison of the forecasting models and the resulting daily CDs. Section \ref{sec:conclusion} concludes the paper with a summary of the findings and discussions.

\section{Data Description}\label{sec:data_exploration}
We have used the NHTS of 2017 \cite{NHTS2017}, which has been widely used to model travel behaviors, as our main data source. This survey includes two data files: one for representing the vehicle information, e.g., fuel type and car models in the sampled households; the other is for the recorded trips made by members in those households. We selected the trips made by EVs by cross referencing the households with hybrid and electrical fuel types from the vehicle data. The data provide rich features of the travel behavior. We will use the following features: house ID, person ID, trip start and end time, trip duration, and trip distance. Table \ref{T:survey_data} shows a portion of the trips with the key features as an example. Additionally, we have selected trips with less than 200 miles trip distances to include more frequent daily trips rather than occasional long-distance trip. Because considering all the recorded trips will add uncommon trips with unreasonably high trip distances, which will hinder the training process of machine learning algorithms.
\begin{table}
\centering
\renewcommand{\arraystretch}{1}
\footnotesize
\captionsetup{labelsep=space,font={footnotesize,sc}}
\caption{Trip chain example}\label{T:survey_data}
\centering
\resizebox{0.9\columnwidth}{!}{
\begin{tabular}{||c|c|c|c|c|c||}
\hline\hline
\multicolumn{1}{|c|}{\multirow{2}{*}{\tabincell{c}{House\\ ID}}} & \multicolumn{1}{c|}{\multirow{2}{*}{{\tabincell{c}{Person  \\ ID}}}} & \multicolumn{1}{c|}{\multirow{2}{*}{{\tabincell{c}{$T^\mathrm{str}$  \\ (h)}}}} & \multicolumn{1}{c|}{\multirow{2}{*}{{\tabincell{c}{$T^\mathrm{end}$  \\ (h)}}}} & \multicolumn{1}{c|}{\multirow{2}{*}{\tabincell{c}{Duration\\ (min)}}} & \multicolumn{1}{c|}{\multirow{2}{*}{\tabincell{c}{Distance\\ (mile)}}}
\\ [2ex] \hline
\multicolumn{1}{|c|}{30000041}   & 1 & 8    & \multicolumn{1}{c|}{9.5}    & \multicolumn{1}{c|}{90}     & \multicolumn{1}{c|}{68.4}        \\

\multicolumn{1}{|c|}{30000041}   & 1   & 18  & \multicolumn{1}{c|}{20}   & \multicolumn{1}{c|}{120}     & \multicolumn{1}{c|}{73.72}        \\

\multicolumn{1}{|c|}{30000041}   & 2   & 7  & \multicolumn{1}{c|}{7.25}   & \multicolumn{1}{c|}{15}     & \multicolumn{1}{c|}{0.68}        \\
\multicolumn{1}{|c|}{30000041}   & 2   & 8  & \multicolumn{1}{c|}{8.25}   & \multicolumn{1}{c|}{15}     & \multicolumn{1}{c|}{0.68}        \\
\hline\hline
                            
\end{tabular}
}
\end{table}

Initially, the raw survey data include 923,572 records of trips, and the vehicle data show 6,416 EVs among the households. After limiting the trips to be less than 200 miles and selecting the EV trips only, 52,094 records remain. Moreover, travel behaviors change over the week especially on the weekends. Therefore, we separated the remaining data to weekdays (WDs) and weekends (WEDs) trips -- 40,917 trips for WDs and 11,240 for WEDs. In the next step, we form the trip chains based on the person and household IDs, meaning that for rows of the trip data, we find the trips made by each person in a household, and order them based on the time of the day. After forming the trip chains, the input features of our forecasting model will be the current trip's start time, end time, duration, and distance. Accordingly, we will forecast the same parameters for the next trip except for the trip duration. As an example, we show the input features and target values of the model in Table \ref{T:inputs_targets} based on the four trips presented in Table \ref{T:survey_data}.
\begin{table}
\centering
\renewcommand{\arraystretch}{0.9}
\footnotesize
\captionsetup{labelsep=space,font={footnotesize,sc}}
\caption{Input and target data example}\label{T:inputs_targets}
\centering
\resizebox{0.9\columnwidth}{!}{
\begin{tabular}{||c|c|c|c|c|c||}
\hline \hline
\multicolumn{3}{|c|}{\multirow{2}{*}{\tabincell{c}{Input features \\ current trip}}} & \multicolumn{3}{c|}{\multirow{2}{*}{\tabincell{c}{Targets \\ next trip}}} \\ [3ex]
\hline
\multicolumn{1}{|c|}{\multirow{2}{*}{\tabincell{c}{$T^\mathrm{str}$  \\ (h)}}} & \multicolumn{1}{c|}{\multirow{2}{*}{{\tabincell{c}{$T^\mathrm{end}$  \\ (h)}}}}  & \multicolumn{1}{c|}{\multirow{2}{*}{{\tabincell{c}{Distance \\ (mile)}}}} & \multicolumn{1}{c|}{\multirow{2}{*}{\tabincell{c}{$T^\mathrm{str}$  \\ (h)}}} & \multicolumn{1}{c|}{\multirow{2}{*}{\tabincell{c}{$T^\mathrm{end}$  \\ (h)}}} & \multicolumn{1}{c|}{\multirow{2}{*}{\tabincell{c}{Distance\\ (mile)}}}
\\ [2.5ex] \hline
\multicolumn{1}{|c|}{8}   & 9.5     & \multicolumn{1}{c|}{68.4}    & \multicolumn{1}{c|}{18}     & \multicolumn{1}{c|}{20}   & \multicolumn{1}{c|}{73.72}     \\

\multicolumn{1}{|c|}{7}   & 7.25    & \multicolumn{1}{c|}{0.68}   & \multicolumn{1}{c|}{8}  &8.25   & \multicolumn{1}{c|}{0.68}        \\

\hline\hline
                            
\end{tabular}
}
\end{table}

\begin{figure*}[htb]
\centering

\begin{tikzpicture}[auto]
  \node[input, name = input] (input){};
  
  \node[blocksina, right = 1.7cm of input,  font = \normalsize] (tw) {EV Trips};
  \node (dot1) [dot,left = 0.001mm of tw]{};
  \node (dot2) [dot,right = 0.001mm of tw]{};
  
  \node[blocksinaa, above  = 0.7cm of input,  font = \large] (tr) {Trip data};
  \node[blocksina, below = 0.7cm of input,  font = \normalsize] (vd) {Vehicle data};
  \node[blocksina, right = 3cm of tr,  font = \normalsize] (weekdays) {Weekdays};
  \node[blocksina, right  = 3 cm of vd,  font = \normalsize] (weeknddays) {Weekends};
  \node[blocksina,  right = 3cm of tw,  font = \normalsize] (tripch) {Trip Chain};
    \node (dottcl) [dot,left = 0.001mm of tripch]{};
  \node (dottcr) [dot,right = 0.001mm of tripch]{};
    \draw [out=0, in=180,  thick]  (weekdays) to (tripch);
  \draw [out=0, in=180,  thick]  (weeknddays) to (tripch);

  \draw [out=0, in=180,  thick]  (tr) to (tw);
  \draw [out=0, in=180,  thick]  (vd) to (tw);
    \draw [out=0, in=180,  thick]  (tw) to (weekdays);
  \draw [out=0, in=180,  thick]  (tw) to (weeknddays);
  \node[blocksina, right = 3.4cm of weekdays,  font = \small] (inputs)  {\framebox{\normalsize
    {\begin{varwidth}{\linewidth}{\,\,\textbf{Input features:} \\ current trip's:\begin{itemize}
        \item Start time
        \item Distance
      \item Duration
      \item End time
     \end{itemize}}\end{varwidth}}
 }};  
  \node[blocksina, below = 0.3cm of inputs,  font = \small] (target) { 
  \framebox{\normalsize
    {\begin{varwidth}{\linewidth}{\,\, \textbf{ Targets:} \\  next trip's: \begin{itemize}
        \item Start time
        \item Distance
      \item End time
     \end{itemize}}\end{varwidth}}
 }}; ;
  
    \draw [out=0, in=180,  thick]  (tripch) to (inputs);
  \draw [out=0, in=180,  thick]  (tripch) to (target);
  
  \node[blocksina, right = 5.5cm of tripch,  font = \normalsize] (preprocess) {Preprocess};
  \draw [out=0, in=180,  thick]  (inputs) to (preprocess);
  \draw [out=0, in=180,  thick]  (target) to (preprocess);
  \node[blocksina, right = 0.8cm of preprocess,  font = \normalsize] (model) {Model};
  \draw[-latex, line width=0.30mm] (preprocess) -- (model);

\end{tikzpicture}

\caption{Data exploration}\label{Fig:data_exploration}
\centering
\end{figure*}
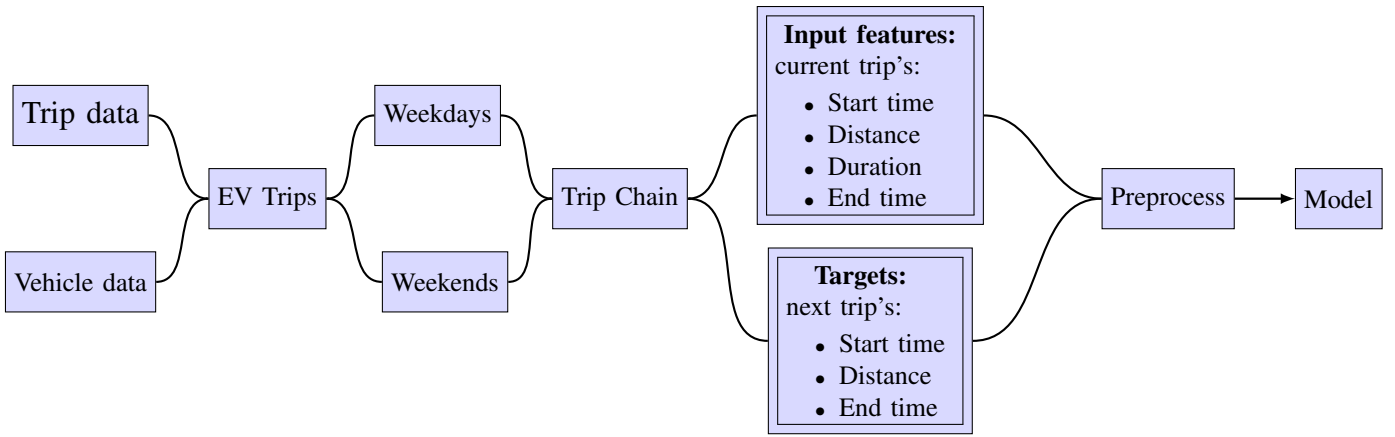

Fig. \ref{Fig:data_exploration} illustrates the summary of the steps described, and in data preprosesing step,  we split the data into training and test sets (75$\%$ training and 25$\%$ test data), and re-scaled the input features based on the training data before feeding them to the forecasting models. 

\section{Charging demand modeling} \label{sec:demand_model}
For modeling the CD of vehicles, we use house ID and vehicle ID to distinguish the trips made by each unique vehicle and calculate its charging status based on the trip distance as follows:
\begin{align}
    \mathrm{SOC}^{\mathrm{arr}}_{v,k} = \mathrm{SOC}^{\mathrm{dep}}_{v,k} - \frac{r_v\cdot d_k}{\mathrm{Cap}^{\mathrm{bat}_v}} \times 100, \label{eq:SOC_calulation}
\end{align}
where $\mathrm{SOC}^{\mathrm{arr}}_{v,k}$ is the state of charge (SOC) of EV $v$ at the end of a trip $k$ and $\mathrm{SOC}^{\mathrm{dep}}_{v,k}$ is the SOC of EV $v$ at the beginning of the trip $k$. $d_k$ represents the distance of the trip and $r_v$ and $\mathrm{Cap}^{\mathrm{bat}}$ are EV's electricity consumption rate (kWh/mile) and battery capacity (kWh), respectively. 

The first/last trip of a day start/end at home locations. To model the charging at locations other than home, we assume that EVs will be able to charge at the end of trips with more than 1 hour dwelling time.  Otherwise, the SOC of the vehicle will be updated based on (\ref{eq:SOC_calulation}). Note that a sensitivity analysis can be done on selecting the acceptable dwelling time for charging the vehicle, but most of the trips either have short dwelling times (less than 1 hour) or very long dwelling times (couple of hours), and considering 1 hour is reasonable. The dwelling time is calculated as the time difference between the end time of a trip and the start time of the next trip. This assumption allows us to model the possibility of charging at work or other viable locations during the day that studies such as \cite{jahangir2019charging} did not consider.

CD ($\mathrm{CD}_{v,t}$) during charging period can be calculated using (\ref{eq:charging_demand}),
\begin{align}
    \mathrm{CD}_{v,t} = \alpha_v \cdot \eta_v \cdot \Delta T, \label{eq:charging_demand}
\end{align}

which depends on charging rate $\alpha_v$, charging efficiency $\eta_v$ and charging duration $\Delta T$. $\Delta T$ is limited to the dwelling time or the required time $T^\mathrm{req}$ for the full charge, which depends on the arrival SOC of the vehicle, as shown in (\ref{eq:t_req}) and (\ref{eq:charging_duration}).
\begin{align}
    T^\mathrm{req} = \frac{(1 - \mathrm{SOC}^{\mathrm{arr}}_v)\cdot \mathrm{Cap}^{\mathrm{bat}_v}}{\alpha_v \cdot \eta_v} \label{eq:t_req}\\
    \Delta T = \mathrm{min}(T^\mathrm{str}_{k+1} - T^\mathrm{end}_k, T^\mathrm{req}) \label{eq:charging_duration}
\end{align}

CD will be considered as load in power system starting from the trip end time ($T^\mathrm{end}_k$) until the end of charging duration ($\Delta T$). After charging, the SOC of the vehicle will be updated based on (\ref{eq:soc_charge}).

\begin{align}
    \mathrm{SOC}^\mathrm{dep}_{v,k+1} = \mathrm{SOC}^{\mathrm{arr}}_{v,k} + \frac{\mathrm{CD}_{v,t}}{\mathrm{Cap}^{\mathrm{bat}_v}}
    \label{eq:soc_charge}
\end{align}

It is evident that the temporal behavior of CD depends on the arrival time of drivers at their destinations $T^\mathrm{end}_k$ and the start time of their next trip $T^\mathrm{str}_{k+1}$. Additionally, the CD value depends on the trip distance $d_k$. These are the same three targets of our forecasting model, and we can estimate the CD of vehicles during the day based on them.
\section{Forecasting models} \label{sec:Models}

We will forecast the three main trip parameters: trip start and end time and travel distance of each trip based on the same parameters of the previous trip. This is the main approach used in most of the studies \cite{li2018gis, jahangir2019charging, mansour2020machine}. However, most of these studies overlook an important feature i.e., the trip duration that, according to our simulations, improves the accuracy of forecasting the travel parameters. Here, we will add this feature to the input data and train both simple and complex machine learning models. The considered simple models are KNN, decision tree (DT), and random forest (RF), and the considered complex models are  ANN-based models including one layer ANN, deep ANN (DANN), recurrent NN (RNN), and long short-term memory (LSTM). Scikit learn library is used to train KNN, DT, and RF, and TensorFlow library is used to train ANN, DANN, RNN, and LSTM. The performance of each forecasting model is based on its training hyper parameters (e.g., number of neighbors for KNN, tree depth for DT and RF, and learning rate for ANN, DANN, RNN, LSTM). We have considered a range of parameters for each model and selected the parameters with the least forecasting error on the test dataset. Root meant square error (RMSE) is used as the error criterion, as formulated in (\ref{eq:RMSE}).
\begin{align}\label{eq:RMSE}
    \mathrm{RMSE} = \sqrt{\sum_{i=1}^n \frac{(\hat{y}_i - y_i)^2}{n}},
\end{align}
where $\hat{y}_i$ and $y_i$ are the estimated and real target values respectively, and $n$ is the number of samples.
The summary of the selected model parameters are presented in Table \ref{T:model_parameters} for WDs and WEDs data and for each forecasting target.

\begin{table}
\centering
\renewcommand{\arraystretch}{1}
\footnotesize
\captionsetup{labelsep=space,font={footnotesize,sc}}
\caption{Forecasting model parameters}\label{T:model_parameters}
\centering
\begin{tabular}{|c|c|c|c|c|c|}
\hline\hline
\multicolumn{2}{|c|}{Target} &\tabincell{c}{KNN \\ neighbors} &\tabincell{c}{DT \\ tree depth} & \tabincell{c}{RF\\ tree depth} & \tabincell{c}{Learning \\ rate} \\ \hline
\multirow{2}{*}{\tabincell{c}{Start\\ time}} & WD   &  17   & 8   &  10  & 1e-3   \\ \cline{2-6} 
                       & WED   &  17   &  5  &  11  & 1e-3  \\ \hline
\multirow{2}{*}{\tabincell{c}{End \\ time}}   & WD  & 15    &  5  &  9  & 1e-2   \\ \cline{2-6} 
                       & WED  &  17   &  5  &  12  &  1e-2  \\ \hline
\multirow{2}{*}{distance}   & WD  &  17   &  6  &  20  & 7e-3    \\ \cline{2-6} 
                       & WED &   14  & 4   &   20 &  7e-5   \\ \hline \hline
\end{tabular}
\end{table}

Training ANN-based models relies on many hyper parameters. Learning rate is one of the important parameters, as presented in Table \ref{T:model_parameters}. The other parameter is the network structure. We tested different structures and the networks reported in Table \ref{T:ANN_structure} give us the least RMSE values.  
\begin{table}
    \centering
\renewcommand{\arraystretch}{1}
\footnotesize
\captionsetup{labelsep=space,font={footnotesize,sc}}
\caption{Ann-based models' network structures with Number of neurons on each layer}\label{T:ANN_structure}
\centering
\resizebox{0.7\columnwidth}{!}{%
\begin{tabular}{|c|c|c|c|c|}
\hline \hline
       \multirow{2}{*}{Layer} & \multirow{2}{*}{ANN} & \multirow{2}{*}{DANN} & \multirow{2}{*}{RNN} & \multirow{2}{*}{LSTM}\\ [1.5ex]
        \hline
         1 & 600 & 600 & 8 & 32\\
         \hline
         2 & - & 300 & 8 & 32 \\
         \hline
         3 & - & 100 & - & -\\ \hline \hline
\end{tabular}
}
\end{table}

\section{Implementation and results}\label{sec:results}

\subsection{Trip parameters}
\begin{figure}
  \centering
  {\includegraphics[width=0.75\linewidth]{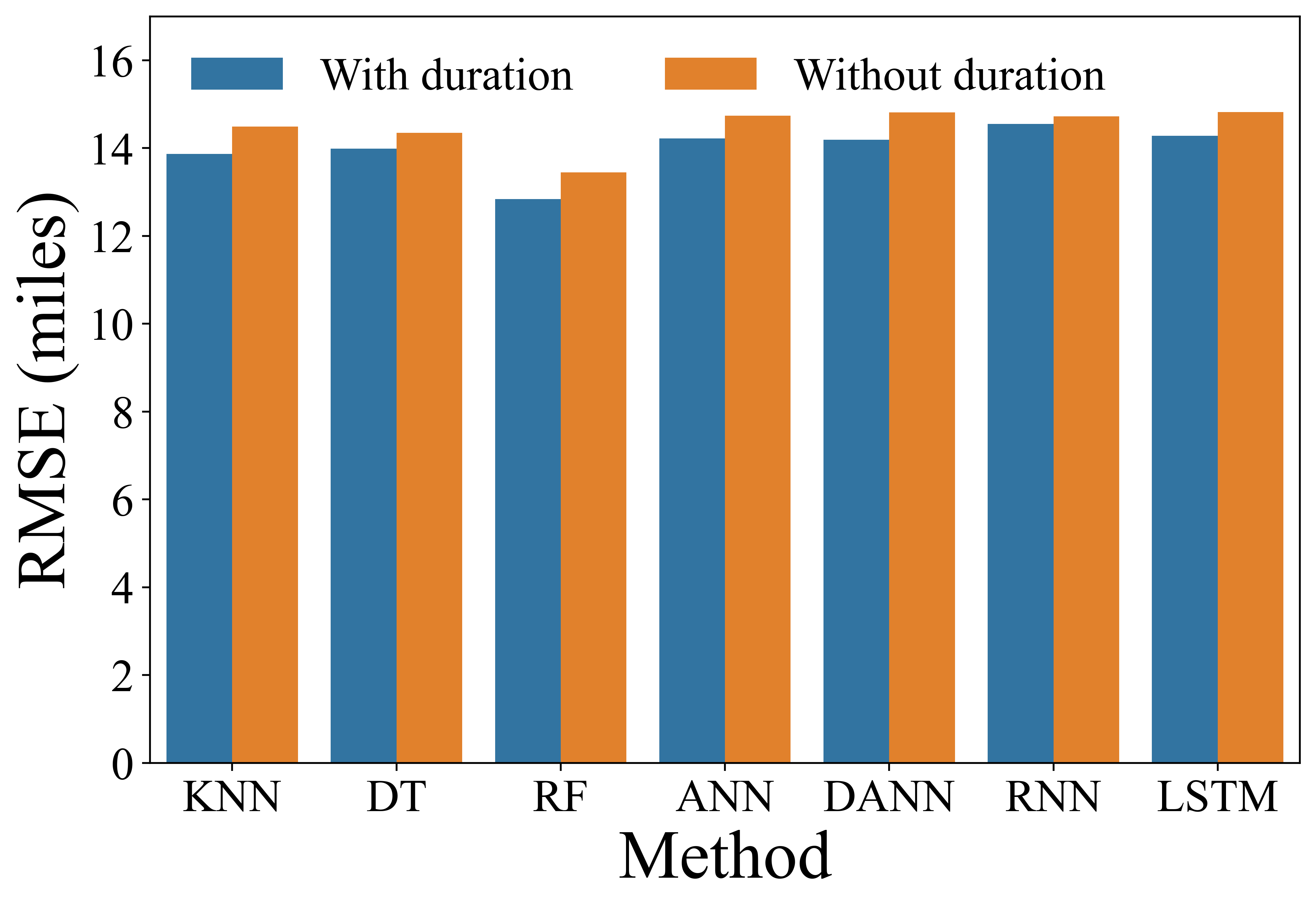}\label{fig:duration}}
    \captionsetup{justification=raggedright,singlelinecheck=false}
  \caption{RMSE value of the foretasted trip distance on WD with and without trip duration as an input feature.}
  
\end{figure}

Based on the estimated models developed in Section \ref{sec:Models}, we foretasted the trip parameters and calculated the RMSE and 95 $\%$ confidence interval (CI) for the real and predicted values on the test data sets. For highlighting the importance of using trip duration as an input feature, we measured the RMSE value of forecasting the trip parameters with and without this feature. Fig. \ref{fig:duration} represents the impact of considering trip duration by comparing the RMSE value for the forecast of trip start time on WDs with and without trip duration as an input feature. It is evident that the forecasts without trip duration have less accuracy (higher RMSE values). Forecasts for other parameters on both day types showed similar different accuracy levels, which indicates the importance of trip duration as an input feature. 

Tables \ref{T:forecasting_RMSE} and \ref{T:forecasting_RMSE_ANN} represent the error values for all the defined trip parameters on WDs and WEDs with incorporating trip duration as an input feature. Additionally, the 95 $\%$ confidence interval (CI) for the input data and the foretasted parameters are reported in parentheses for each method. We expected to achieve less accuracy using the simple machine learning models such as KNN and DT. However, the results show that these models outperformed more sophisticated ANN-based models in terms of RMSE value. The RF model provided better accuracy for almost every target parameter except for the start time on WDs (see Table \ref{T:forecasting_RMSE}). 

Comparing the 95 $\%$ CIs indicate that the ANN-based methods provide forecasts with less standard deviation compared to the simple methods. In other words, the ranges of intervals for the simple methods are larger than ANN-based methods, which are closer to the ranges of intervals for the input data. This is another indicator of the better performance of simple methods compared to the ANN-based methods.  

Even though ANN-based methods had less accuracy in terms of RMSE values, we can not discredit their performance because extracting the trend of travel behavior during the day is more important in this study. The results of the CD estimation in the next section will further elaborate this statement.
\begin{table}
\centering
\footnotesize
\captionsetup{labelsep=space,font={footnotesize,sc}}
\caption{Forecasting errors (RMSE) and 95 $\%$ CIs for simple models}
\centering
\resizebox{0.9\columnwidth}{!}{%
\begin{tabular}{|c|c|c|c|c|}
\hline \hline
\multicolumn{2}{|c|}{\multirow{2}{*}{Target}} &\multirow{2}{*}{\tabincell{c}{KNN}} &\multirow{2}{*}{\tabincell{c}{DT}} & \multirow{2}{*}{\tabincell{c}{RF}} \\ [2ex] \hline
\multirow{2}{*}{\tabincell{c}{Start\\ time (h)}} & {\tabincell{c}{WD \\ (14.85,15.02)}}   &  {\tabincell{c}{2.34\\ (14.85,14.99)}}   & {\tabincell{c}{2.35\\ (14.84,14.98)}}   &  {\tabincell{c}{2.32\\ (14.85,14.99)}}  \\ \cline{2-5} 
                       & {\tabincell{c}{WED \\ (14.65,14.96)}}   &  {\tabincell{c}{1.96\\ (14.75,15)}}   &  {\tabincell{c}{1.98\\ (14.76,15.02)}}  &  {\tabincell{c}{1.92\\ (14.76,15.03)}} \\ \hline
\multirow{2}{*}{\tabincell{c}{End \\ time (h)}}   & \tabincell{c}{WD \\ (15.17,15.34)}   & \tabincell{c}{2.44 \\ (15.17,15.31)}     & \tabincell{c}{2.43 \\ (15.16,15.29)} & \tabincell{c}{2.37 \\ (15.16,15.29)} \\ \cline{2-5} 
& \tabincell{c}{WED \\ (14.96,15.27)}  & \tabincell{c}{2.04 \\ (15.05,15.31)}   & \tabincell{c}{1.99 \\ (15.07,15.32)}  & \tabincell{c}{1.91 \\ (15.05,15.31)} \\ \hline
\multirow{2}{*}{\tabincell{c}{Distance\\ (mile)}}   & \tabincell{c}{WD \\ (7.59,8.29)}   &  \tabincell{c}{13.87 \\ (7.67,7.99)}    & \tabincell{c}{13.99 \\ (7.65,8)}  & \tabincell{c}{12.54 \\ (8.05,8.5)}  \\ \cline{2-5} 
                       &  \tabincell{c}{WED \\ (8.11,9.67)}  & \tabincell{c}{19.9 \\ (8.74,9)}   & \tabincell{c}{16.85 \\ (8.39,9.11)}   & \tabincell{c}{12.61 \\ (8.85,9.93)} \\ \hline \hline
\end{tabular}\label{T:forecasting_RMSE}
}
\end{table}

\begin{table}
\centering
\renewcommand{\arraystretch}{1}
\footnotesize
\captionsetup{labelsep=space,font={footnotesize,sc}}
\caption{Forecasting errors (RMSE) and 95 $\%$ CIs for ANN-based models}
\centering
\resizebox{\columnwidth}{!}{%
\begin{tabular}{|c|c|c|c|c|c|}
\hline \hline
\multicolumn{2}{|c|}{\multirow{2}{*}{Target}} &  \multirow{2}{*}{\tabincell{c}{ANN}} & \multirow{2}{*}{DANN} & \multirow{2}{*}{RNN} & \multirow{2}{*}{LSTM}\\ [2ex] \hline
\multirow{2}{*}{\tabincell{c}{Start\\ time (h)}} & \tabincell{c}{WD \\ (14.85,15.02)}  & \tabincell{c}{2.39 \\ (14.41,14.55)}  & \tabincell{c}{2.35 \\ (14.23,14.37)} & \tabincell{c}{2.45 \\ (14.37,14.52)} & \tabincell{c}{2.37 \\ (14.38,14.53)} \\ \cline{2-6} 
                       & \tabincell{c}{WED \\ (14.65,14.96)} & \tabincell{c}{2.04 \\ (14.28,14.56)} & \tabincell{c}{2.06 \\ (14.32,14.61)} & \tabincell{c}{1.99 \\ (14.34,14.61)} & \tabincell{c}{2.01 \\ (14.45,14.73)}\\ \hline
\multirow{2}{*}{\tabincell{c}{End \\ time (h)}}   & \tabincell{c}{WD \\ (15.17,15.34)}  & \tabincell{c}{2.46 \\ (14.74,14.88)} & \tabincell{c}{2.84 \\ (14.65,14.8)} & \tabincell{c}{2.45 \\ (14.88,15.01)} & \tabincell{c}{2.46 \\ (14.8,14.95)} \\ \cline{2-6} 
                       & \tabincell{c}{WED \\ (14.96,15.27)}  & \tabincell{c}{2.06 \\ (14.74,15.02)} & \tabincell{c}{2.05 \\ (14.78,15.06)} & \tabincell{c}{2.03 \\ (14.81,15.08)} & \tabincell{c}{2.06 \\ (14.78,15.05)}\\ \hline
\multirow{2}{*}{\tabincell{c}{Distance\\ (mile)}}   &  \tabincell{c}{WD \\ (7.59,8.29)} & \tabincell{c}{14.22 \\ (4.87,5.14)} & \tabincell{c}{14.19 \\ (5.34,5.66)} & \tabincell{c}{14.55 \\ (4.65,4.94)} & \tabincell{c}{14.28 \\ (5.13,5.4)}  \\ \cline{2-6} 
                       & \tabincell{c}{WED \\ (8.11,9.67)}  &  \tabincell{c}{17.47 \\ (5.11,5.64)} & \tabincell{c}{17.59 \\ (5.34,6.08)} & \tabincell{c}{17.96 \\ (5.76,6.37)}  & \tabincell{c}{17.9 \\ (4.98,5.51)} \\ \hline \hline
\end{tabular}\label{T:forecasting_RMSE_ANN}
}
\end{table}

\subsection{Charging demand pattern}
Using the formulation presented in Section \ref{sec:demand_model}, we can derive the charging pattern of vehicles based on the trips made by each vehicle of the households from the survey data. We will consider Nissan Leaf to be the primary EV model in the fleet as the most popular EV reported in Alternative Fuels Data Center \cite{AFDC2020} with the following characteristics: charging rate $\alpha = 6.6$ (kWh), battery capacity $\mathrm{Cap^\mathrm{bat}} = 30$ (kWh), charging efficiency $\eta = 90\%$, and consumption rate $r = 0.15$ (kWh/mile) \cite{AFDC2020}. It is important to note that the charging rate $\alpha$ can incorporate both fast and slow charging modes by considering higher or lower rates. In this paper, we have considered the charging rate based on the battery characteristics of the vehicle to maintain generality. 

\begin{figure}
  \centering
  \subfloat[]{\includegraphics[width=0.85\linewidth]{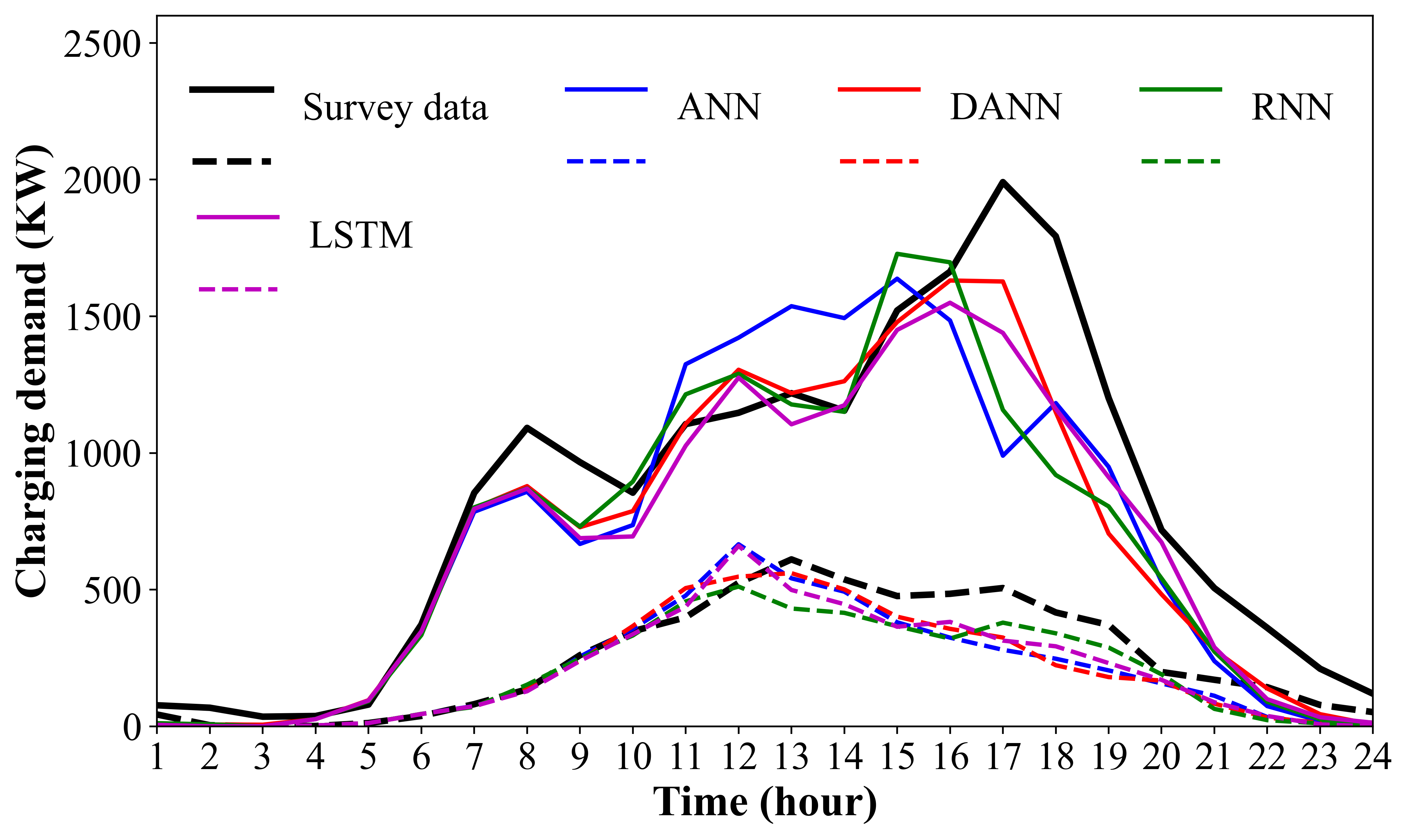}\label{fig:ANN-based}}
  
  \subfloat[]{\includegraphics[width=0.85\linewidth]{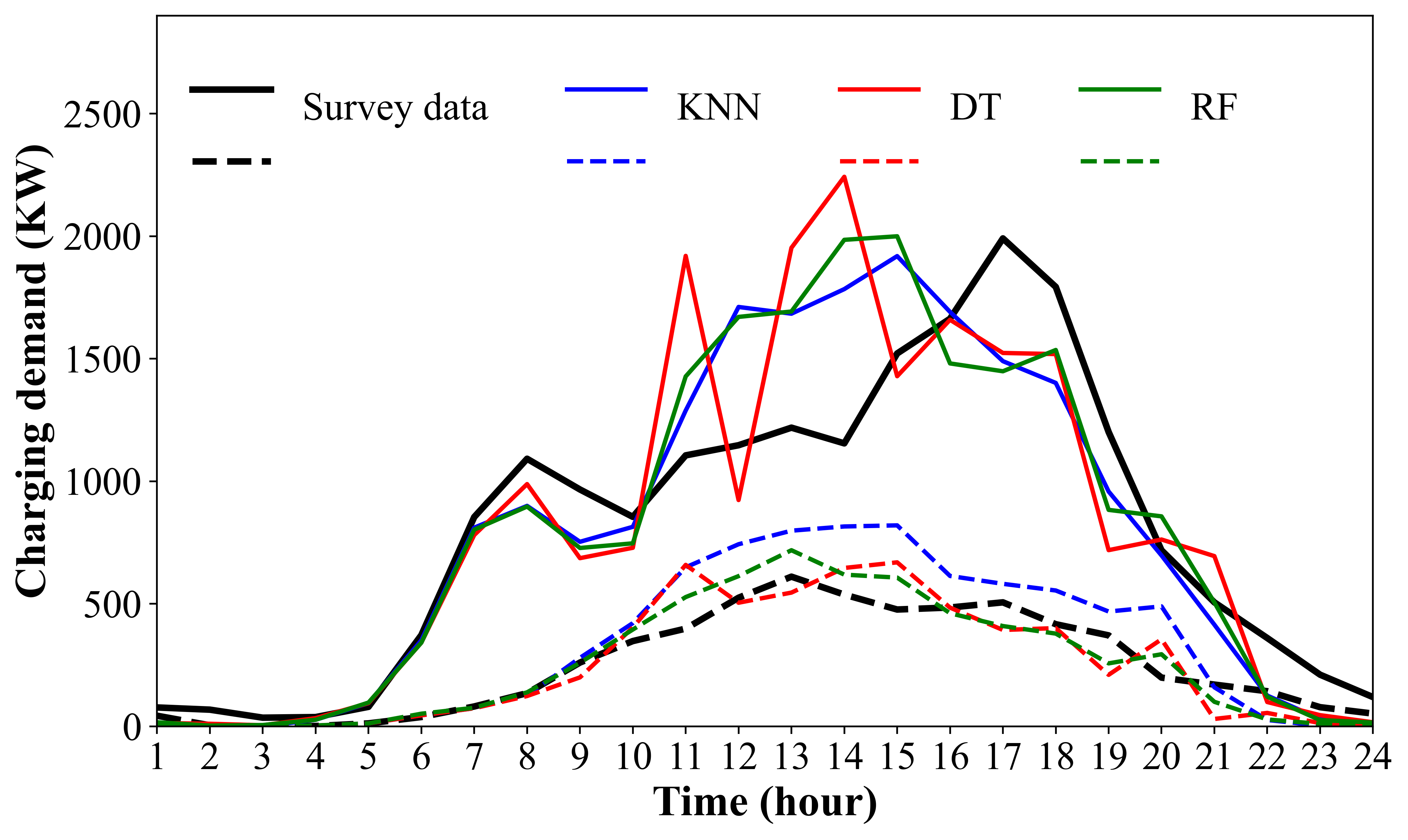}\label{fig:simples}}
    \captionsetup{justification=raggedright,singlelinecheck=false}
  \caption{Estimated daily charge demand for (a) ANN-based modes (b) Simpler models. Solid lines and dashed lines represent WD and WED, respectively.}
  
\end{figure}

We calculated the daily CD based on the predicted travel parameters on the test data covering 13,023 trips of nearly 6,500 users and their corresponding data on the survey data to further investigate the performance of the models. The,  forecasting errors in predicting the parameters will cause shifts in both CD and charging time because the CD depends on the driving distance; and the start time and duration of charging depend on drivers' trip start time and end time. We present the estimated daily CD with two figures, one for the ANN-based approaches (Fig. \ref{fig:ANN-based}) and the other for the simpler models (Fig. \ref{fig:simples}). The CD for WDs are shown with solid lines and WEDs are shown with dashed lines. Same colors are used for the same models. 

Even though ANN-based model had higher error values in forecasting each driving parameters, they were able to predict the daily CD pattern more accurately (see Fig. \ref{fig:ANN-based}) compared to the simpler models such as KNN or DT (see Fig. \ref{fig:simples}). We observe more accurate results on the WEDs, where there are no significant spikes. Among the models, the RF model provides the best accuracy in terms of RMSE, and the DANN model better predicts the temporal pattern of the CD during the day. ANN-based models follow the pattern of the CD better because these methods train more modeling parameters compared to the simpler methods, and they try to adapt to the trend of the data rather than fitting to the exact target values. 

\section{Conclusion and discussion}\label{sec:conclusion}
Travel behavior of EV owners plays an important role in estimating EVs' daily charging demand (CD). Travel behavior of EVs can be analyzed with trip parameters i.e., trip distance and trip start and end time, and the temporal behavior of EVs' CD can be estimated based on these parameters. Moreover, an accurate CD estimation can help power system operators to plan and allocate enough energy sources to provide the required energy during the day. In this paper, we present a data-driven approach to predict the trip parameters of EVs based on NHTS survey data using different machine learning techniques and calculated the resultant CD during the day. The simulation results show better accuracy in terms of trip start time and end time compared to the study done on the same dataset \cite{jahangir2019charging} and provide a more realistic daily CD compared to the same study. Additionally, We found that simple machine learning techniques such as KNN, DT, and RF can provide acceptable prediction for driving parameters. However, they were not able to capture the daily temporal pattern of CD, while the ANN-based methods perform better in this task.

The CD forecasting can be further developed into day ahead demand forecasting by generating daily trip chains based on the historical trip data, and the presented work can be applied to estimate the daily CD. The spatial distribution of CD can also be modeled based on the destination of trip chains.   

\bibliographystyle{./bibliography/IEEEtran}

\bibliography{./bibliography/my_library.bib}

\end{document}